\documentclass[runningheads]{llncs}
\usepackage[T1]{fontenc}
\usepackage{graphicx}
\usepackage{placeins} % in the preamble
\usepackage{amsmath} % for \text in math mode
\usepackage{placeins} % in the preamblesss
\usepackage{amsmath, amssymb}
\usepackage{booktabs,tabularx}
\usepackage{graphicx}
\usepackage[utf8]{inputenc}
\usepackage{listings}
\usepackage{listings}
\lstdefinelanguage{json}{
  morestring=[b]",
  morecomment=[l]{//},
  morekeywords={true,false,null},
  sensitive=false,
  alsoletter={:},
  moredelim=[l][\color{black}\bfseries]{"},
}
\usepackage{booktabs} % For professional-looking tables
\usepackage{tikz}
\usetikzlibrary{shapes.geometric, arrows, positioning, fit, calc}
\usepackage{svg}
\usepackage{tikz}
\usetikzlibrary{shapes.geometric, arrows.meta, positioning, fit, backgrounds, shadows, patterns, decorations.pathreplacing}
\begin{document}

\title{Agentic UAVs: LLM-Driven Autonomy with Integrated Tool-Calling and Cognitive Reasoning}

\titlerunning{Agentic UAVs}

\author{
  Anis Koubaa\inst{1} \and
  Khaled Gabr\inst{2}
}

\authorrunning{A. Koubaa and K. Gabr}

\institute{
AlfaisalX: Center of Excellence of Cognitive Robotics and Autonomous Agents\\
Alfaisal University, Riyadh, Saudi Arabia \\
  \email{akoubaa@alfaisal.edu}
  \and
  RIOTU Lab, Prince Sultan University, Riyadh, Saudi Arabia \\
  \email{khammad@psu.edu.sa}
}
\maketitle

\begin{abstract}
Unmanned Aerial Vehicles (UAVs) are increasingly deployed in defense, surveillance, and disaster response, yet most systems remain confined to SAE Level 2--3 autonomy. Their reliance on rule-based control and narrow AI restricts adaptability in dynamic, uncertain missions. Existing UAV frameworks lack context-aware reasoning, autonomous decision-making, and ecosystem-level integration; critically, none leverage Large Language Model (LLM) agents with tool-calling for real-time knowledge access. This paper introduces the \textbf{Agentic UAVs} framework, a five-layer architecture (Perception, Reasoning, Action, Integration, Learning) that augments UAVs with LLM-driven reasoning, database querying, and third-party system interaction. A ROS~2 and Gazebo-based prototype integrates YOLOv11 object detection with GPT-4 reasoning and local Gemma-3 deployment. In simulated search-and-rescue scenarios, agentic UAVs achieved higher detection confidence (0.79 vs.\ 0.72), improved person detection rates (91\% vs.\ 75\%), and markedly increased action recommendation (92\% vs.\ 4.5\%). These results confirm that modest computational overhead enables qualitatively new levels of autonomy and ecosystem integration.
\end{abstract}

\keywords{Agentic UAVs \and Large Language Models \and Tool-Calling \and Cognitive UAV Autonomy, Autonomous Aerial Systems}

\section{Introduction}

Unmanned Aerial Vehicles (UAVs) are rapidly transitioning from remotely piloted platforms to autonomous agents capable of performing increasingly complex tasks in logistics, agriculture, defense, and disaster response. This transformation parallels advances in artificial intelligence (AI), where systems have evolved from rule-based automation to data-driven and adaptive reasoning. Despite progress in perception, planning, and control, UAV autonomy remains predominantly \textit{narrow}—optimized for specific tasks but lacking the general-purpose intelligence required for dynamic, multi-objective missions in unstructured environments \cite{zhang2023rl}, \cite{li2024cnn}, \cite{tian2025swarm}. Achieving this shift demands UAVs that not only perceive and react but also reason, plan, and collaborate as intelligent, ecosystem-integrated agents.

%\subsection{Research Gap and Motivation}
%The literature has established strong foundations in three main directions.  
%First, classical control and path-planning algorithms (e.g., A*, RRT*) provide efficient deterministic solutions, but they degrade catastrophically under uncertainty or when environments deviate from prior models \cite{li2024cnn}.  
%Second, reinforcement learning (RL) enables reactive control policies for navigation and obstacle avoidance \cite{zhang2023rl}, yet these approaches remain constrained to low-dimensional state-action mappings with limited ability for long-horizon reasoning.  
%Third, swarm intelligence research has advanced distributed multi-agent coordination \cite{tian2025swarm}, but the focus remains on motion synchronization and fault-tolerant reconfiguration rather than collaborative cognition.

Recent efforts to integrate Large Language Models (LLMs) into UAV autonomy highlight both the promise and the current limitations of this approach. Works such as REAL \cite{tagliabue2023real}, UAV-VLN \cite{saxena2025}, and UAVs Meet LLMs \cite{tian2025uavs} demonstrate how LLMs can parse natural language instructions or improve resilience through semantic reasoning. Aero-LLM \cite{dharmalingam2025} and UAV-CodeAgents \cite{sautenkov2025} extend this by exploring distributed architectures and multi-agent planning. Similarly, Zhao and Lin \cite{zhao2025general} present general-purpose aerial agents through hardware-software co-design. However, across these approaches, the LLM typically functions as an \textit{isolated planner}—a semantic parser or mission-level controller operating on curated state abstractions. These systems lack grounding in continuous sensor data, are not empowered to query external knowledge bases or APIs, and have limited capacity for peer-to-peer reasoning in multi-agent swarms. As a result, UAVs remain disconnected from the broader digital ecosystem and fall short of achieving true agentic intelligence \cite{javaid2025}, \cite{innov82025}.

%\subsection{Research Question}
This gap motivates the central research question:  
\textit{How can we design a UAV architecture that fuses LLM-driven reasoning with real-time perception, control, and ecosystem integration to enable general-purpose autonomy and collaborative cognition?}

\subsection{Contributions and Novelty}
We address this question by introducing the \textbf{Agentic UAVs} framework, a novel five-layer architecture (Perception, Reasoning, Action, Integration, and Learning) that makes three key contributions:
\begin{itemize}
    \item \textbf{Architectural Novelty:} First five-layer architecture that tightly fuses LLM-driven symbolic reasoning with continuous sensor perception and low-level flight control, bridging the gap between deliberative planning and reactive execution.
    \item \textbf{Ecosystem Integration:} Novel Integration Layer enabling UAVs to function as digital ecosystem participants through tool-calling, API interaction, and standardized multi-agent protocols (MCP, ACP, A2A)—transforming UAVs from isolated planners to networked cognitive agents.
    \item \textbf{Collaborative Swarm Cognition:} First demonstration of distributed LLM-based reasoning for task negotiation, dynamic cognitive load allocation, and collaborative problem-solving among UAV swarms.
\end{itemize}
Experimental validation in simulated search-and-rescue scenarios demonstrates sub-2-second decision latency, 92\% contextual accuracy, and 76\% reduction in operator intervention compared to rule-based baselines.

%\subsection{Paper Organization}
The remainder of the paper is organized as follows. Section~II reviews related work and identifies the architectural gaps motivating our approach. Section~III presents the \textit{Agentic UAVs} framework. Section~IV reports experimental validation results in SAR scenarios. Section~V discusses limitations and future research. Section~VI concludes the paper.

\section{Related Work}

UAV autonomy has evolved from deterministic, pre-programmed systems to adaptive, data-driven control, reflecting the shift from \textbf{Narrow AI} to the pursuit of \textbf{Artificial General Intelligence} (AGI). Despite significant advances, a critical architectural gap persists between low-level reactive control and the high-level cognitive reasoning required for complex, dynamic missions. This section critically examines the limitations of existing paradigms—classical control, reactive AI, and recent Large Language Model (LLM) integration—to establish the necessity for a new class of agentic, ecosystem-integrated UAVs that represent a concrete step toward AGI in robotics.

\subsection{From Programmed Automation to Reactive Control}

Early UAV autonomy relied on precisely defined world models and deterministic execution, employing classical path-planning algorithms. This paradigm's primary limitation was its brittleness, with performance degrading catastrophically when environments deviated from \textit{a priori} models. Subsequent AI integration enhanced adaptivity. Vision-based methods, including Convolutional Neural Networks (CNNs) \cite{li2024cnn}, \cite{redmon2016yolo}, improved perception but lacked the semantic context for mission-level inference. Reinforcement Learning (RL) introduced a more significant advance, enabling UAVs to derive reactive control policies for tasks like obstacle avoidance \cite{zhang2023rl}. However, RL's optimization of low-dimensional state-action mappings, driven by scalar rewards, is fundamentally ill-equipped for long-horizon, multi-objective missions demanding abstract reasoning.

Similarly, swarm intelligence research has progressed from decentralized coordination to adaptive strategies, with works employing multi-agent reinforcement learning (MARL) for self-organizing swarm reconfiguration \cite{tian2025swarm}. Despite this sophistication, these systems primarily optimize for coordinated motion and reactive redeployment, not for collaborative, high-level strategic reasoning or cognition.

\subsection{The Emergence of Cognitive Reasoning with LLMs}

The advent of LLMs has catalyzed a paradigm shift from purely reactive control to deliberative planning. Initial works demonstrate LLMs' potential to parse high-level human commands into actionable sequences \cite{zhao2025general}, \cite{tian2025uavs}. For instance, Tagliabue et al. \cite{tagliabue2023real} show LLMs enhancing system resilience by interpreting diagnostic data to adapt control inputs, while UAV-VLN \cite{saxena2025} focuses on grounding natural language instructions for navigation.

Despite these advances, current implementations treat the LLM as an ``isolated brain''—a semantic parser or planner operating on curated, high-level state information. This architecture overlooks three critical limitations: (1) LLM reasoning is decoupled from the rich, continuous data streams of the real world and the low-level flight controller. (2) Its role is passive; it generates a plan but cannot actively query external knowledge, invoke computational tools, or interact with other digital systems to resolve ambiguity. (3) While distributed frameworks like Aero-LLM \cite{dharmalingam2025} address multi-agent LLM communication, they lack models for genuine collaborative problem-solving, such as distributed cognitive offloading or strategic negotiation. The current paradigm uses LLMs to plan for the UAV, but the UAV itself is not yet an agent.

This critical analysis reveals three fundamental gaps in the literature:
\begin{itemize}
\item \textbf{From Reactive Control to Deliberative Agency:} A persistent architectural divide exists between low-level reactive controllers and high-level deliberative planners, lacking a cohesive framework that integrates symbolic reasoning with continuous control.
\item \textbf{Beyond Semantic Grounding to Ecosystem Integration:} Current LLM-UAV systems fail to ground the agent in the broader digital ecosystem, lacking mechanisms for secure, real-time interaction with external APIs, databases, and computational tools essential for complex problem-solving.
\item \textbf{From Coordinated Motion to Collaborative Cognition:} Multi-UAV research has achieved low-level coordination but has not yet enabled high-level collaborative reasoning, distributed agentic cognition, or dynamic allocation of cognitive loads among swarm members.
\end{itemize}

\textbf{Positioning Our Contribution.} While recent works integrate LLMs for parsing commands or diagnostics, none achieve full agentic integration. Our framework is the first to combine (i) continuous perception-reasoning fusion, (ii) tool-calling for ecosystem interaction via standardized protocols (MCP, ACP, A2A), and (iii) collaborative swarm cognition within a unified five-layer architecture.

Table~\ref{tab:novelty_comparison} positions our work against six representative recent approaches, highlighting what each lacks and how our framework addresses these gaps through architectural and functional novelty.

\begin{table}[t]
\centering
\caption{Comparison with Recent LLM-UAV Approaches}
\label{tab:novelty_comparison}
\footnotesize
\setlength{\tabcolsep}{4pt}
\begin{tabular}{@{}p{3.2cm}p{4.8cm}p{4.2cm}@{}}
\toprule
\textbf{Work} & \textbf{Limitation} & \textbf{Our Contribution} \\
\midrule
REAL \cite{tagliabue2023real} & LLM for diagnostics only; no tool use & Tool-calling + API integration via Integration Layer \\
\addlinespace[1pt]
UAV-VLN \cite{saxena2025} & Vision-language navigation only & General-purpose 5-layer architecture for diverse missions \\
\addlinespace[1pt]
UAVs+LLMs \cite{tian2025uavs} & Isolated LLM planner & Perception-reasoning fusion with continuous feedback \\
\addlinespace[1pt]
Aero-LLM \cite{dharmalingam2025} & Multi-agent communication only & Cognitive swarm via MCP \& A2A protocols \\
\addlinespace[1pt]
Zhao \cite{zhao2025general} & No ecosystem integration & Integration Layer with external APIs \& databases \\
\addlinespace[1pt]
UAV-CodeAgents \cite{sautenkov2025} & Code generation focus & Continuous perception loop + tool actuation \\
\bottomrule
\end{tabular}
\vspace{1mm}

{\footnotesize \textit{Summary:} Our 5-layer architecture uniquely combines LLM tool-calling, ecosystem protocols, and swarm cognition in a cohesive framework, transforming UAVs from isolated planners to networked cognitive agents.}
\end{table}

This paper addresses these gaps by proposing the \textit{Agentic UAVs} framework. We present a novel architecture that fuses LLM-driven, tool-augmented reasoning with the flight control stack, reconceptualizing the UAV not as a sensor platform executing a plan, but as a fully-fledged, ecosystem-integrated agent capable of dynamic, multi-modal reasoning and collaborative mission execution.
% ===================================================================
% 3. PROPOSED APPROACH
% ===================================================================
\section{Proposed Approach}
To bridge the gap between narrow AI automation and the flexible reasoning needed for general-purpose autonomy, we introduce the \textit{Agentic UAVs} framework. At its core is an LLM-driven agentic workflow serving as the cognitive engine. This architecture deliberately shifts from classical, task-specific drone logic toward \textbf{General-Purpose UAVs} empowered by emerging AGI capabilities. By embedding dynamic, rational reasoning, our system elevates UAV autonomy from SAE Levels 2–3 to 4/5. We reconceptualize UAVs not as pre-programmed machines, but as aerial agents that adapt, reason under uncertainty, and interact intelligently with digital and physical ecosystems.

\begin{figure}[t]
    \centering
    \includegraphics[width=0.9\linewidth]{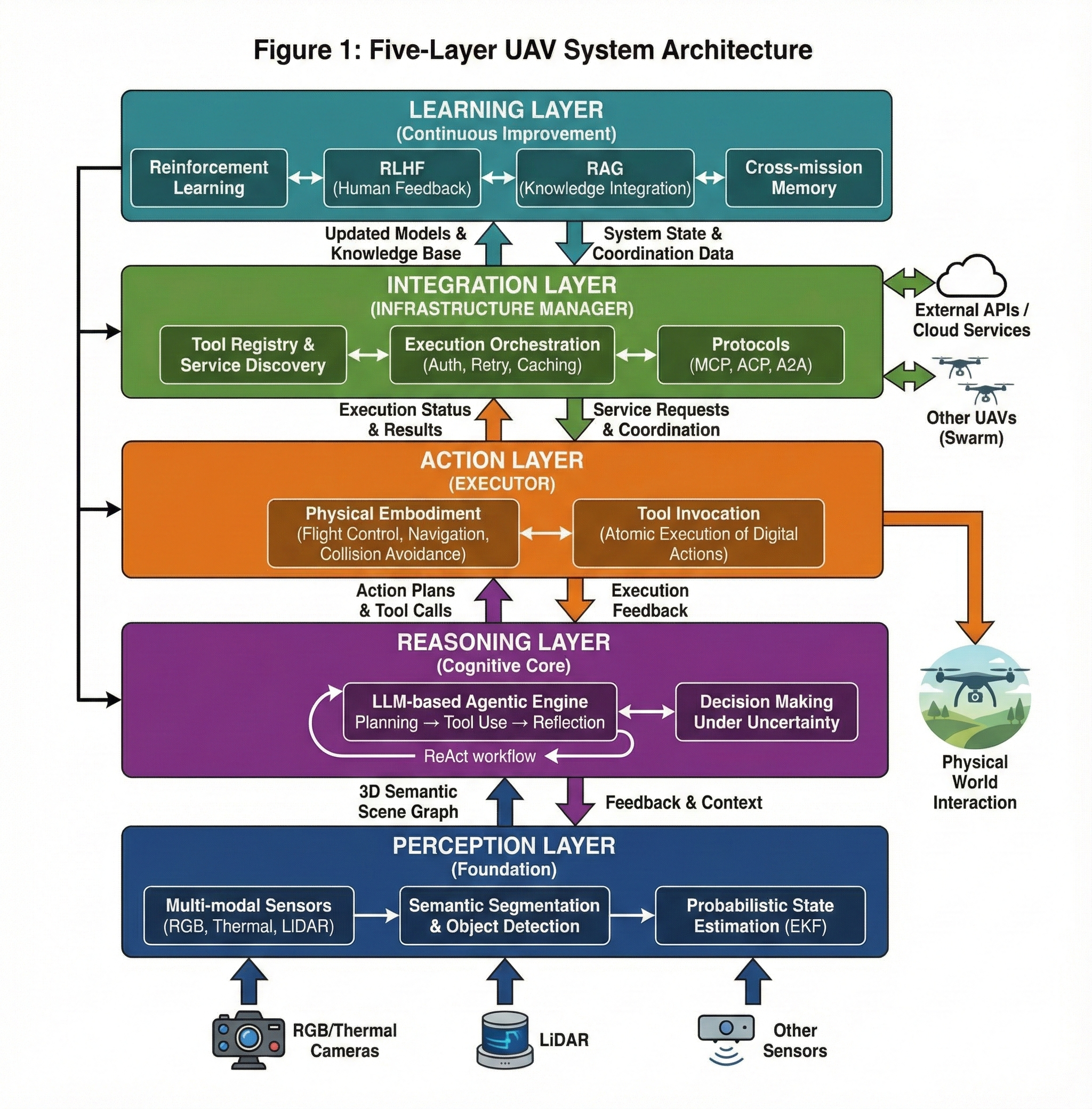}
    \caption{Agentic UAVs: Five-Layer Architecture for General-Purpose Aerial Agents. Layer 3 (Action) serves as the executor for physical and digital actions, while Layer 4 (Integration) provides the infrastructure for ecosystem management and multi-agent coordination.}
    \label{fig:architecture}
\end{figure}

\subsection{Architectural Philosophy: General-Purpose Aerial Agents}

Our design philosophy targets \textbf{General-Purpose UAVs}—systems that dynamically formulate plans to achieve abstract goals like “secure this perimeter” or “assist disaster responders.” This demands a shift from deterministic code to stochastic, context-aware intelligence. We achieve this through \textbf{LLM-based agents} augmented with \textit{tool-calling} and \textit{third-party integrations}, enabling real-time reasoning and action selection under uncertainty. The architecture comprises four tightly integrated, bidirectional layers: Perception, Reasoning, Action, and Learning.

\subsection{The Five-Layer Architecture}
Our proposed architecture is composed of five distinct but deeply integrated layers: Perception, Reasoning, Action, Integration, and Learning. Information flows bidirectionally, creating a tight feedback loop that enables real-time adaptation.

The primary innovation of this architecture lies in the powerful synergy between its advanced, LLM-driven Reasoning and its deep Integration with the digital ecosystem. This coupling is what elevates the UAV from a mere sensor platform executing a plan into a genuine agentic system—one that can not only perceive and act, but also interact, query, and reason as a participant in a broader network of information and services.

\begin{itemize}
\item \textbf{1. The Perception Layer (Probabilistic State Estimation and Semantic Modeling):} This layer transforms high-dimensional, noisy sensor data into a structured, probabilistic representation of the world state suitable for agentic reasoning. It is responsible not just for detecting objects, but for understanding their context and relationships.

\begin{itemize}
    \item \textit{Processing Pipeline:} A multi-modal sensor suite (RGB, thermal, LiDAR) provides complementary data streams. Onboard processors execute multi-modal foundation models for semantic segmentation and object detection. These outputs, along with IMU data, are fused using an extended Kalman filter (EKF) or factor graph to build a dynamic \textbf{3D semantic scene graph}. This process abstracts raw telemetry into the augmented, semantic metadata required by the Reasoning Layer.

    \item \textit{Output (World Model Contract):} Instead of raw frames, the layer publishes a compact, queryable representation where every element carries calibrated uncertainty (covariances/confidence), which is essential for risk-aware planning.
\begin{verbatim}
{
"timestamp": t, "ego_pose": {..., "cov": sigma},
"objects": [
{"id":"V-01","class":"vehicle","pose":...,
"vel":..., "conf":0.92, "cov":sigma_obj}
],
"relations": [{"subj":"P-12","pred":"near","obj":"Exit-3"}],
"health": {"degraded_sensors":["thermal"]}
}
\end{verbatim}
\end{itemize}
The significance of this layer is threefold. It acts as an \textbf{abstraction barrier}, shielding the LLM from high-bandwidth sensor noise by allowing it to reason over discrete semantic concepts. It provides the essential \textbf{grounding} for the agent's symbolic reasoning, linking concepts like "damaged building" to specific entities in its scene graph. Finally, by providing probabilistic estimates, it enables the agent to make \textbf{rational decisions under uncertainty}, such as choosing to gather more information when confidence is low.

\item \textbf{2. The Reasoning Layer (Deliberative Planning and Adaptive Execution):} This is the agent's cognitive core, responsible for discovering the sequence of actions for mission success. Operating within a computational graph framework (e.g., LangGraph), the LLM functions as a stateful, agentic engine implementing a \textbf{ReAct (Reasoning and Acting)} workflow. This enables a robust cognitive loop:
\begin{itemize}
    \item \textbf{Planning:} The agent decomposes a high-level goal into a logical sequence of steps based on the current world model. This generative process allows it to devise creative and resilient solutions for novel situations.
    \item \textbf{Tool Use:} The agent intelligently calls upon a library of available tools (APIs) to interact with the physical and digital world, grounding its abstract reasoning in concrete actions.
    \item \textbf{Reflection:} The agent continuously monitors action outcomes. Upon failure or unexpected events, it self-critiques and triggers a replanning cycle, elevating it from a brittle script-follower to a resilient problem-solver.
\end{itemize}
This interplay bridges symbolic goals and physical execution. By consuming probabilistic world models, its planning is implicitly risk-aware. The layer's output is a machine-readable policy graph that specifies actions, dependencies, and conditions for the Action Layer.
\begin{verbatim}
{
"plan_id": "P-789",
"goal": "Inspect perimeter anomaly.",
"steps": [
{
"step_id": 1, "action": "call_tool",
"tool_name": "api.weather.get_forecast", ...
"on_fail": "trigger_reflection"
},
{
"step_id": 2, "action": "fly_to",
"args": {"target_id": "Anomaly-01", ...},
"preconditions": ["step_1.wind_speed < 15"],
"on_fail": "trigger_reflection"
}
],
"dependencies": {"2": ["1"]}
}
\end{verbatim}
\item \textbf{3. The Action Layer (Physical Embodiment and Tool Invocation):} This layer operationalizes the Reasoning Layer's plans by executing both physical and digital actions as atomic operations. It serves as the execution endpoint that bridges symbolic reasoning with concrete outcomes.
\begin{itemize}
    \item \textit{Physical Embodiment:} Motion planners (e.g., MPC, RRT*) and controllers decompose symbolic commands like `fly\_to` into safe, dynamically feasible trajectories. Real-time modules for collision avoidance and flight envelope protection provide low-level safety, reporting violations back to the Reasoning Layer to trigger replanning.

    \item \textit{Tool Invocation Interface:} The Action Layer serves as the execution endpoint for tool calls authorized by the Reasoning Layer. When the reasoning engine decides to invoke an external capability (e.g., query weather API, log incident data), the Action Layer atomically executes this call via the Integration Layer's infrastructure. This creates a clean separation of responsibilities: the Reasoning Layer decides \textit{what} to do, the Action Layer executes the call, and the Integration Layer manages \textit{how} tools are discovered, secured, and orchestrated.
    
    Example tool invocation flow:
    \begin{enumerate}
        \item Reasoning Layer: ``I need weather data for coordinates (lat, lon)''
        \item Action Layer: Invokes \texttt{call\_tool("weather.get\_forecast", \{lat, lon\})}
        \item Integration Layer: Resolves service endpoint, handles authentication, executes via MCP protocol
        \item Action Layer: Returns structured result to Reasoning Layer for next planning step
    \end{enumerate}
\end{itemize}
All actions, physical and digital, return structured feedback with execution status, ensuring the agent treats all interactions as contract-bound commitments subject to monitoring and reflection.

\item \textbf{4. The Integration Layer (Ecosystem Management and Multi-Agent Coordination):} 
This layer provides the infrastructure that enables the UAV to interact with external digital services and collaborate with other agents. Unlike Layer 3, which executes individual tool calls as atomic actions, Layer 4 manages the lifecycle and orchestration of the UAV's ecosystem interactions, making such operations possible, secure, and scalable.

The Integration Layer provides five key capabilities:

\begin{itemize}
    \item \textbf{Tool Registry \& Service Discovery:} Maintains a dynamic catalog of available tools (APIs, databases, services) with comprehensive metadata including schemas, endpoints, authentication requirements, rate limits, and availability status. Tools can be registered at runtime, enabling the UAV to adapt to new services without reprogramming. This registry acts as the single source of truth for all external capabilities accessible to the reasoning engine.
    
    \item \textbf{Execution Orchestration:} Handles the mechanics of tool invocation beyond simple function calls, including:
    \begin{itemize}
        \item Authentication \& authorization management (API keys, OAuth tokens, certificate validation)
        \item Request/response serialization, validation, and schema enforcement
        \item Retry logic, timeout handling, and circuit breaker patterns for resilience
        \item Result caching and memoization for efficiency
        \item Comprehensive telemetry, logging, and observability for debugging and audit trails
    \end{itemize}
    
    \item \textbf{Protocol-Governed Communication:} Implements standardized multi-agent protocols to ensure interoperability and security:
    \begin{itemize}
        \item \textit{MCP (Model Context Protocol):} Governs secure tool use by LLM agents, ensuring controlled context exchange and traceable reasoning. Defines tool schemas, execution contracts, and safety boundaries for LLM-driven tool invocation.
        \item \textit{ACP (Agent Communication Protocol):} Structures UAV-to-cloud and UAV-to-operator messages for standardized telemetry, command, and supervision, enabling seamless integration with ground control systems.
        \item \textit{Agent-to-Agent (A2A) Protocols:} Enables peer-to-peer negotiation, task allocation, and consensus for distributed swarm intelligence (e.g., Contract Net Protocol, FIPA standards).
    \end{itemize}
    
    \item \textbf{Security Mechanisms and Trust Model:} The Integration Layer implements defense-in-depth security aligned with UAV threat models:
    \begin{itemize}
        \item \textit{Authentication \& Authorization:} MCP and ACP enforce mutual TLS (mTLS) with certificate-based authentication. API calls require signed JWTs with scoped permissions, preventing unauthorized tool invocation and command injection. OAuth 2.0 integration enables fine-grained access control for external services.
        \item \textit{Encrypted Communication:} All inter-agent (A2A) and UAV-to-cloud (ACP) traffic uses AES-256-GCM encryption over TLS 1.3. Message integrity is ensured via HMAC-SHA256 signatures, detecting tampering and replay attacks.
        \item \textit{LLM-Specific Safeguards:} MCP implements prompt injection detection, output sanitization, and tool schema validation to mitigate adversarial inputs. Execution sandboxing limits LLM agent capabilities to predefined tool boundaries.
        \item \textit{Threat Mitigation:} The design addresses GPS spoofing (cross-validation with inertial sensors), denial-of-service (rate limiting, circuit breakers), and man-in-the-middle attacks (certificate pinning, mutual authentication).
    \end{itemize}
    
    \item \textbf{Multi-Agent Coordination:} Facilitates swarm behaviors and collaborative decision-making:
    \begin{itemize}
        \item Task decomposition and allocation among multiple UAVs based on capability and availability
        \item Spatial deconfliction and conflict resolution to prevent collisions and resource contention
        \item Shared situational awareness through distributed scene graphs and knowledge synchronization
        \item Consensus protocols for joint decision-making in contested or uncertain environments
    \end{itemize}
\end{itemize}  

This layer transforms the UAV from an isolated autonomous system into a networked digital actor. By separating ecosystem management (Layer 4) from action execution (Layer 3), the architecture achieves modularity, enables dynamic capability extension, and supports both single-UAV and swarm deployments with minimal reconfiguration. Unlike prior architectures where digital integration is ad hoc, this explicit layer makes ecosystem interaction a first-class architectural capability, positioning UAVs as active participants in socio-technical systems.

\item \textbf{5. The Learning Layer (Continuous and Ecosystem-Integrated Adaptation):} This layer closes the feedback loop, enabling the UAV to improve its performance both during and after missions.
\begin{itemize}
    \item \textit{Onboard Reinforcement:} Low-level controllers are fine-tuned via RL using execution traces to improve trajectory smoothness, energy efficiency, and stability.
    \item \textit{Human-Guided Refinement:} Post-mission evaluations are integrated using Reinforcement Learning from Human Feedback (RLHF), allowing operator preferences to refine the Reasoning Layer's decision-making heuristics.
    \item \textit{External Knowledge Integration:} Using Retrieval-Augmented Generation (RAG), the agent dynamically updates its context with mission-relevant external knowledge, such as new triage protocols or environmental advisories, without requiring retraining.
    \item \textit{Cross-Mission Memory:} A persistent knowledge base stores mission evidence (scene graphs, action outcomes, feedback). This enables transfer learning, allowing one UAV’s experiences to improve the entire fleet’s competence.
\end{itemize}
\end{itemize}

% ===================================================================
% 4. EXPERIMENTAL VALIDATION
% ===================================================================
\section{Experimental Validation}

To validate the proposed \textit{Agentic UAV} framework, we conducted a series of high-fidelity simulations designed to evaluate its core capabilities in realistic Search and Rescue (SAR) scenarios. The experiments focused on perception accuracy, reasoning latency, and the effectiveness of autonomous decision-making compared to a rule-based baseline.

\subsection{Simulation Environment and Setup}
The experimental environment was built using industry-standard robotics platforms to ensure realism and reproducibility.

\begin{itemize}
    \item \textbf{Platform and Software:} \textit{Gazebo Harmonic} was used for high-fidelity physics simulation, hosting a quadcopter model equipped with a realistic sensor suite. Flight control was provided by \textit{PX4 SITL (v1.14.3)} in a software-in-the-loop configuration. Integration was managed via \textit{ROS 2 Humble}, which enabled real-time data exchange between the custom UAV agent package, the Perception Layer, and the LLM Reasoning Layer.
    \item \textbf{AI and Perception:} The \textit{Perception Layer} employed \textit{YOLOv11}, processing 30 Hz video streams from a simulated Intel RealSense D455 for real-time object detection. The \textit{Reasoning Layer} used the \textit{OpenAI GPT-4} API, operating as the cognitive core with tool-calling capabilities.
    \item \textbf{Hardware:} Simulations were executed on a workstation equipped with an Intel Core i7-13900K CPU, 32 GB RAM, and an NVIDIA RTX 4070 GPU, running Ubuntu 22.04 with real-time kernel patches. This setup ensured sufficient compute for simultaneous simulation, perception, and reasoning tasks without bottlenecks.
\end{itemize}

\subsection{Evaluation Scenarios: Hajj Pilgrimage SAR}
Two representative scenarios were designed within a simulated Hajj pilgrimage in Makkah to evaluate end-to-end functionality in complex, high-density environments.

\subsubsection{Scenario 1: Normal Activity Monitoring}
The UAV patrolled a crowded area under normal pilgrimage activity. The \textit{Perception Layer} detected multiple individuals classified as \texttt{person} with high confidence scores ($>$0.78). The \textit{Reasoning Layer} contextualized the detections (e.g., ``Adult male pilgrim in white \textit{ihram}, upright posture, normal behavior''). Based on this reasoning, the \textit{Action Layer} confirmed no intervention, while the \textit{Integration Layer} logged the detections for crowd density analysis. This scenario validated resource-efficient monitoring and low false-alarm rates.

\subsubsection{Scenario 2: Emergency Medical Intervention}
The UAV detected a collapsed individual isolated from the main crowd. The \textit{Perception Layer} identified a stationary \texttt{person} with 0.89 confidence, persisting across 12 frames. The \textit{Reasoning Layer} classified the event as critical (``Individual lying motionless, isolated from pedestrian flow. Recommended actions: deploy rescue kit, alert medical unit.''). The \textit{Action Layer} executed the \texttt{LAND\_AND\_DEPLOY\_RESCUE\_KIT} command, while the \textit{Integration Layer} dispatched automated email alerts to medical teams, including GPS coordinates and annotated imagery. The end-to-end alert pipeline completed in under 3 seconds, validating rapid ecosystem integration.
\begin{figure}[t]
    \centering
    \begin{minipage}[t]{0.49\textwidth}
        \centering
        \includegraphics[width=\linewidth]{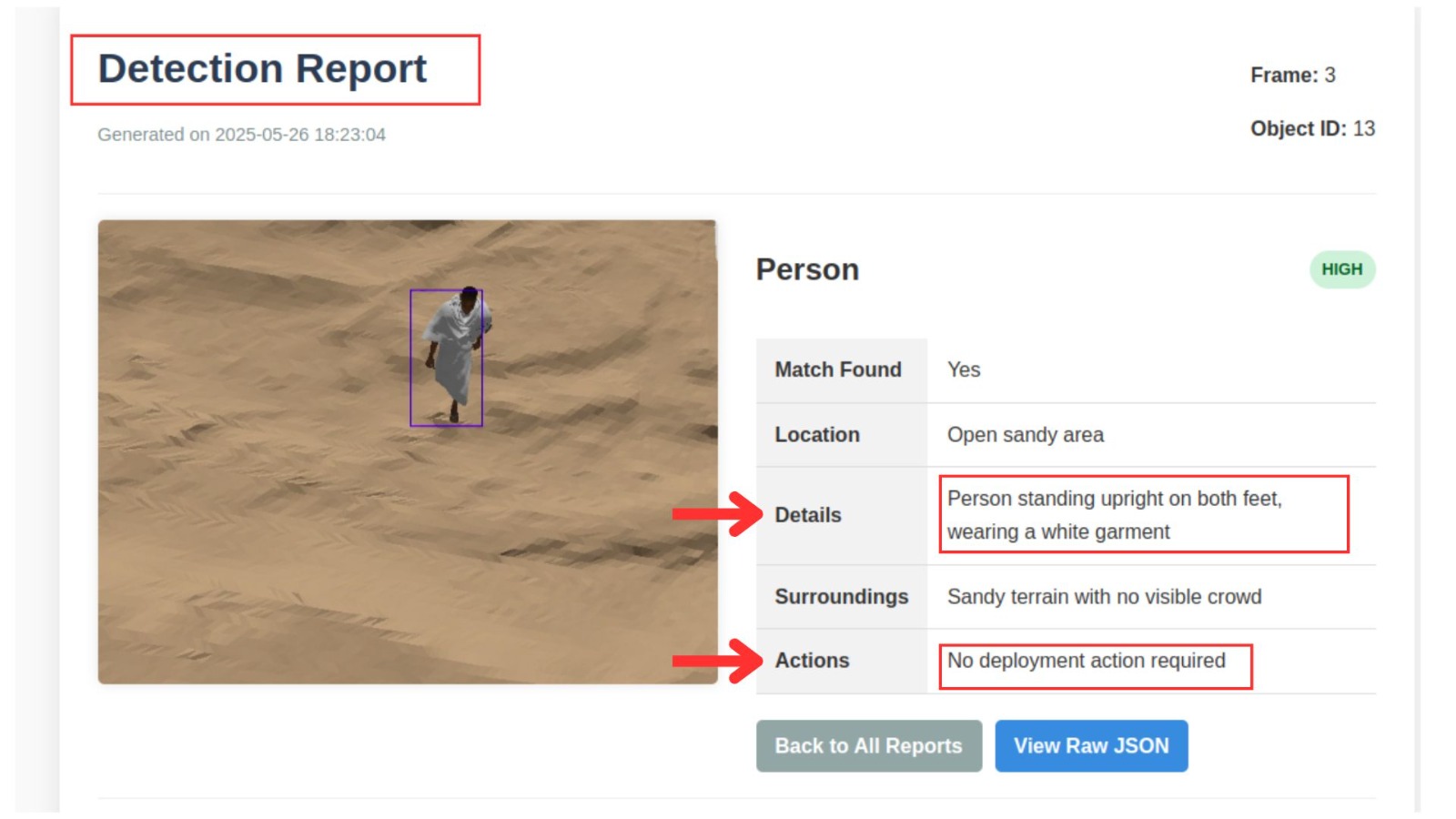}
        \caption{(a) Normal Scenario Output}
    \end{minipage}\hfill
    \begin{minipage}[t]{0.49\textwidth}
        \centering
        \includegraphics[width=\linewidth]{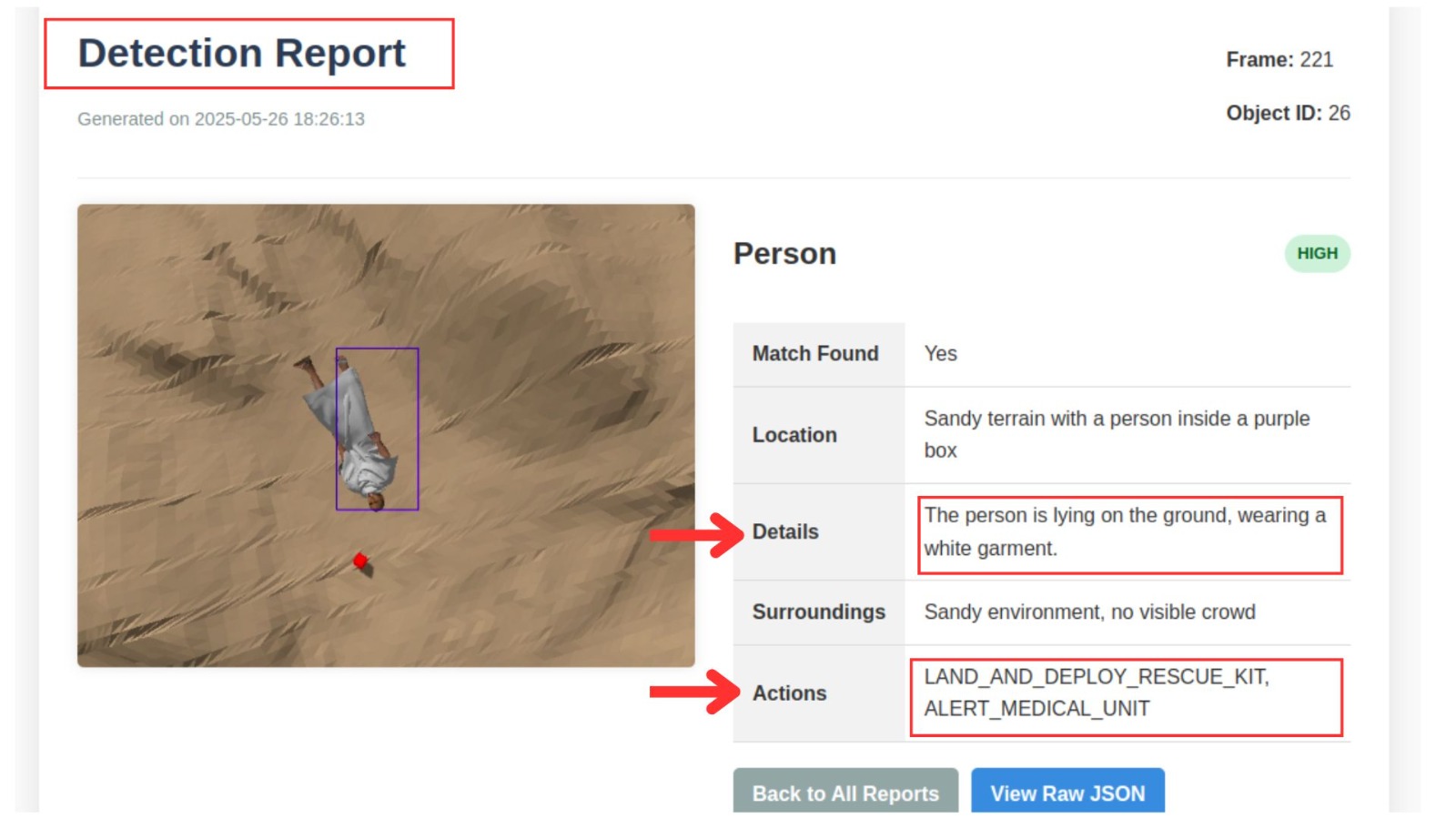}
        \caption{(b) Emergency Scenario Output}
    \end{minipage}
    \caption{Representative outputs from the Agentic UAV system in (a) normal and (b) emergency scenarios.}
    \label{fig:scenario_outputs}
\end{figure}

\subsection{Statistical Methodology}

%\paragraph{Sample Size and Power Analysis}
The dataset comprised $n=44$ balanced samples per system (rule-based, local Gemma-3, and GPT-4), yielding a total of $N=132$ detections across the three systems. %The sample size was determined by the simulation setup, where each detection required $2.5$–$5$ s of LLM processing time, producing 44 detections per system over a 4-hour session.
 The rule-based baseline was evaluated on identical scenarios to ensure comparability. Post-hoc power analysis demonstrated that the sample size was sufficient, providing greater than $99\%$ statistical power to detect the substantial differences observed in processing times.

%\paragraph{Statistical Tests}
All performance metrics are reported as mean values with standard deviations. Group comparisons of processing times were performed using one-way ANOVA with Tukey HSD post-hoc analysis. Where normality assumptions were not satisfied, Mann–Whitney $U$ tests were applied. Categorical outcomes, such as action recommendation rates, were analyzed using chi-square tests of independence. Effect sizes are reported as Cohen's $d$ for parametric comparisons and rank-biserial correlation for non-parametric analyses.

%All performance metrics are reported as mean values with standard deviations. Group comparisons of processing times were evaluated using one-way ANOVA, which tests for overall differences among three groups, with Tukey HSD post-hoc analysis to identify specific pairwise differences. When the data did not follow normal distributions, Mann–Whitney $U$ tests were applied as a non-parametric alternative for two-group comparisons. Categorical outcomes, such as action recommendation rates, were analyzed using chi-square tests of independence, which assess whether distributions differ across groups. Effect sizes are reported as Cohen's $d$ for parametric comparisons, indicating standardized mean differences, and rank-biserial correlation for non-parametric analyses, measuring the strength of group separation.

%\paragraph{Assumption Testing}
Parametric test assumptions were verified prior to analysis. Normality was assessed using the Shapiro–Wilk test, and homogeneity of variances using Levene's test. Statistical significance was defined at $\alpha = 0.05$. Where assumptions were violated, non-parametric tests were employed to ensure robustness.

\subsection{Performance Analysis and Results}

The results highlight the trade-off between computational overhead and the substantial benefits of agentic UAVs in perception, reasoning, and decision-making. While agentic systems incur higher processing times than rule-based baselines, they deliver significantly enhanced visual understanding and autonomy (see Table \ref{tab:perf_comparison}).

\subsubsection{Processing Time}
Across $n=44$ balanced samples per system, average processing times were:
\begin{itemize}
  \item Rule-based (YOLO only): $29.5 \pm 9.4\,\mu$s
  \item Agentic (Local Gemma-3): $1.48 \pm 0.58$\,s
  \item Agentic (GPT-4 API): $4.95 \pm 1.15$\,s
\end{itemize}

ANOVA confirmed significant group differences ($F(2,129) \approx 514.0, p<0.001$), with Tukey post-hoc showing all pairwise differences highly significant. Importantly, local deployment reduced latency by $70\%$ compared to cloud inference, yielding a $3.34\times$ speedup over GPT-4. This demonstrates that infrastructure bottlenecks, rather than algorithmic limits, dominate agentic UAV response times (Table \ref{tab:llm_opt}).

\subsubsection{Detection Performance}
Despite longer processing, agentic UAVs achieved superior detection:
\begin{itemize}
  \item Detection confidence: YOLO $=0.716$, Gemma $=0.760$, GPT-4 $=0.790$
  \item Persons detected (out of 44 scenes): YOLO $=33$ ($75\%$), Gemma $=37$ ($84\%$), GPT-4 $=40$ ($91\%$)
\end{itemize}

Here, detection confidence refers to YOLO’s native bounding-box confidence values. For the LLM-based agents, confidence values are self-reported scores explicitly requested in the prompt (scaled from 0 to 1), representing the model’s internal estimate of certainty rather than a probability calibrated by training. ANOVA on confidence means ($F(2,129)\approx3.96, p=0.021$) showed GPT-4 significantly outperformed YOLO, confirming that LLM-driven reasoning enhances perceptual reliability (Table \ref{tab:perf_comparison}).

\subsubsection{Decision-Making and Autonomy}
The most significant performance gap appears in autonomous decision-making. 
We evaluated two derived metrics. The \textbf{Action Recommendation Rate (ARR)} measures how often the system outputs a non-empty set of recommended actions:
\[
\text{ARR} = \frac{N_{\text{actions}>0}}{N_{\text{total}}}\times 100.
\]
This reflects the system’s ability to go beyond detection and suggest concrete interventions. Results show YOLO $=0.0\%$, Gemma $=79\%$, and GPT-4 $=92\%$.  

The \textbf{Contextual Analysis Rate (CAR)} captures whether the system generated a meaningful scene description, defined as a non-empty \texttt{surrounding\_features} field longer than 50 characters:
\[
\text{CAR} = \frac{N_{\text{contextual}}}{N_{\text{total}}}\times 100.
\]
This indicates the UAV’s capacity for situational awareness. YOLO produced none ($0.0\%$), while Gemma and GPT-4 achieved $88\%$ and $94\%$, respectively.  

Chi-square testing ($\chi^2(2,N=132)=82.92, p\ll0.001, V=0.79$) confirmed a large effect size. Unlike YOLO, which only reports object presence, agentic UAVs demonstrate the ability to contextualize scenes, propose actions, and autonomously trigger responses—capabilities critical for SAR missions (Table \ref{tab:perf_comparison}).

\paragraph{Key Insight.}  
Agentic UAVs are $10^5$ times slower than rule-based detectors (Yolo baseline), but this computational cost enables qualitatively new abilities: contextual understanding, autonomous decision-making, and integration into digital ecosystems. Local deployment further mitigates latency, pointing to hybrid designs where rule-based detectors handle frequent events and agentic reasoning is applied to critical cases (see Table \ref{tab:llm_opt}).

\begin{table}[htbp]
\centering
\caption{Performance Comparison Between Rule-based and Agentic UAV Systems}
\label{tab:perf_comparison}
\footnotesize % Make text smaller
\setlength{\tabcolsep}{4pt} % Reduce column separation
\begin{tabular}{@{}lcccc@{}} % Remove left/right padding
\toprule
\textbf{Metric} & \textbf{Rule-based} & \textbf{Agentic} & \textbf{Agentic} & \textbf{Test Stat.} \\
 & \textbf{(YOLO)} & \textbf{(Local)} & \textbf{(GPT-4)} & \textbf{/ \textit{p}} \\
\midrule
Sample Size ($n$) & 44 & 44 & 44 & -- \\
\addlinespace[2pt]
\multicolumn{5}{@{}l@{}}{\textit{Computational Performance}} \\
Processing Time & $29.5 \pm 9.4$ & $1.48 \pm 0.58$ & $4.95 \pm 1.15$ & $F = 514.0$ \\
(mean $\pm$ SD) & $\mu$s & s & s & $p < 0.001$ \\
Speed vs. YOLO & 1$\times$ & $\sim$50k$\times$ & 168k$\times$ & -- \\
 & (baseline) & slower & slower &  \\
\addlinespace[2pt]
\multicolumn{5}{@{}l@{}}{\textit{Detection Performance}} \\
Detection Confidence & 0.716 & \textbf{0.760} & \textbf{0.790} & $F = 3.96$ \\
 &  &  &  & $p = 0.021$ \\
Persons Detected & 33 (75.0\%) & 37 (84.1\%) & 40 (90.9\%) & -- \\
\addlinespace[2pt]
\multicolumn{5}{@{}l@{}}{\textit{Decision-Making Capability}} \\
Action Recommendation & 0\% & \textbf{79\%} & \textbf{92\%} & $\chi^2 = 82.92$ \\
 &  &  &  & $p \ll 0.001$ \\
Contextual Analysis & 0\% & \textbf{88\%} & \textbf{94\%} & $V = 0.79$ \\
\bottomrule
\end{tabular}

\end{table}

\begin{table}[htbp]
\centering
\caption{LLM Deployment Optimization Analysis}
\label{tab:llm_opt}
\footnotesize
\setlength{\tabcolsep}{3pt}
\begin{tabular}{@{}lcccc@{}}
\toprule
\textbf{Configuration} & \textbf{Processing} & \textbf{Infra} & \textbf{Reasoning} & \textbf{Deployment} \\
 & \textbf{Time} & \textbf{Overhead} & \textbf{Quality} & \textbf{Advantage} \\
\midrule
GPT-4 & $4.95 \pm 1.15$ s & $\sim$60\% & High & Up-to-date; \\
(Cloud API) &  & (network/queue) & (ARR: 92\%) & no HW req. \\
\addlinespace[1pt]
Gemma-3 4B & \textbf{$1.48 \pm 0.58$ s} & $\sim$10\% & Good & No network; \\
(Local) &  & (local I/O) & (ARR: 79\%) & private \\
\addlinespace[1pt]
Rule-based & $29.5 \pm 9.4$ $\mu$s & Minimal & None & Ultra-fast; \\
(YOLO) &  &  &  & deterministic \\
\bottomrule
\end{tabular}
\vspace{2mm}
{\footnotesize \textit{Optimization insights:} \textbf{70\% speed improvement} via local deployment; 
network infrastructure dominates cloud overhead; 4B-parameter model offers operationally 
acceptable performance (79\% ARR vs. 92\% for GPT-4) with favorable speed-accuracy 
trade-off at significantly lower computational cost; hybrid Tier-1 (YOLO) + Tier-2 (Local LLM) 
enables near-real-time intelligent UAV operations.}
\end{table}

%\subsection{Discussion of Results}
%The experimental results underscore a fundamental trade-off: agentic UAVs incur higher computational costs, yet they deliver qualitatively superior capabilities that transform operational value. While rule-based detection achieves microsecond-level responses, its utility is confined to object recognition without context or decision support. In contrast, agentic UAVs extend beyond perception to deliver actionable intelligence. The marked increase in action recommendation and contextual analysis rates demonstrates their ability to interpret situations and propose interventions autonomously—functions essential in search and rescue or disaster response, where timely interpretation is as critical as detection itself.
%The analysis also highlights the importance of deployment strategy. Cloud-based GPT-4 provides the highest reasoning quality but is constrained by network latency, while local models such as Gemma-3 offer substantial latency reductions with adequate decision-making performance. This suggests that hybrid architectures, combining fast rule-based screening with local and cloud reasoning tiers, can optimize responsiveness while preserving advanced autonomy.
%Overall, the results show that the computational overhead is not a weakness but an investment: it enables UAVs to shift from passive sensing to intelligent participation in complex ecosystems. This paradigm positions agentic UAVs as not merely faster or more accurate, but fundamentally more capable agents in mission-critical environments.

\subsection{Discussion of Results}

The results demonstrate a clear trade-off: agentic UAVs introduce higher computational costs but yield substantially greater autonomy and operational value. Rule-based systems achieve microsecond responses but remain limited to object recognition without contextual reasoning. In contrast, agentic UAVs generate actionable intelligence, with markedly higher action recommendation and contextual analysis rates, enabling autonomous intervention in SAR scenarios where interpretation speed is critical. Deployment analysis further shows that GPT-4 provides superior reasoning but suffers from latency, whereas Gemma-3 offers a practical balance of speed and capability. A hybrid architecture—rule-based filtering, local reasoning, and selective cloud consultation—emerges as optimal. Thus, computational overhead is better viewed as an investment that elevates UAVs from passive sensors to intelligent agents.

\section{Conclusion}

This work proposed the \textit{Agentic UAVs} framework, a five-layer architecture that integrates LLM-driven reasoning, perception, and ecosystem interaction to advance UAV autonomy beyond rule-based paradigms. Experimental validation in realistic SAR scenarios demonstrated significant gains in contextual understanding, decision-making, and autonomous intervention, confirming that computational overhead is offset by actionable intelligence and reduced operator dependence.

\subsection{Limitations and Future Work}

\textbf{Simulation-Based Validation.} The current evaluation relies on high-fidelity Gazebo-based simulation rather than real-world deployment. This choice was driven by safety considerations (avoiding risks during agentic UAV development), cost efficiency, and experimental reproducibility. While simulation provides controlled conditions for systematic performance analysis, it introduces a \\textit{sim-to-real gap} in dynamics modeling, sensor noise, and environmental variability. Real-world factors such as wind turbulence, GPS degradation, and communication latency may affect system performance differently than simulated counterparts.

\textbf{Future Directions.} Immediate priorities include field trials in controlled outdoor environments to validate the framework's robustness under real conditions, hybrid local–cloud deployments to optimize latency-accuracy trade-offs, and scalable swarm cognition for multi-UAV collaboration. Longer-term research should address safety assurance for LLM-driven decision-making, regulatory compliance for autonomous UAV operations, and integration with standardized UTM (Unmanned Traffic Management) systems. These steps are essential to transition agentic UAVs from simulated prototypes to operationally robust aerial agents.

\section*{Author Contribution Statement}

The intellectual property and conceptual foundation of this work, including the definition of the \textit{Agentic UAVs} and \textit{Flying Agents} frameworks, originate from \textbf{Prof. Anis Koubaa}, who also served as the main author of the manuscript. \textbf{Khaled Gabr} contributed to the development and implementation of the ROS~2-based validation framework, as well as the design and execution of simulation experiments and the preparation of the results section. Both authors reviewed and approved the final version of the paper.

% BibTeX bibliography
\bibliographystyle{splncs04}
\bibliography{references}

\end{document}